\documentclass[10pt,twocolumn,letterpaper]{article}

\usepackage{cvpr}              

\newif\ifreview
\reviewfalse 

\definecolor{cvprblue}{rgb}{0.21,0.49,0.74}
\usepackage[pagebackref,breaklinks,colorlinks,allcolors=cvprblue]{hyperref}
\usepackage{multirow}
\usepackage{afterpage}

\def\paperID{*****} 
\def\confName{CVPR}
\def\confYear{2025}

\title{AthletePose3D: A Benchmark Dataset for 3D Human Pose Estimation and Kinematic Validation in Athletic Movements}

\author{
Calvin Yeung$^1$ 
\text{  } \text{  } 
Tomohiro Suzuki$^1$
\text{  }\text{  }
Ryota Tanaka$^1$ 
\text{  }\text{  }
Zhuoer Yin$^1$ 
\text{  }\text{  }
Keisuke Fujii$^{1,2}$\\
$^1$ Graduate School of Informatics, Nagoya University, Nagoya, Japan\\
$^2$ Center for Advanced Intelligence Project, RIKEN, Osaka, Japan\\
{\tt\small \{yeung.chikwong,suzuki.tomohiro,tanaka.ryota,yin.zhuoer,fujii\}@g.sp.m.is.nagoya-u.ac.jp}
}

\begin{document}

\twocolumn[{%
\renewcommand\twocolumn[1][]{#1}%
\maketitle
\begin{center}
    \centering
    \captionsetup{type=figure}
    \includegraphics[width=1\linewidth]{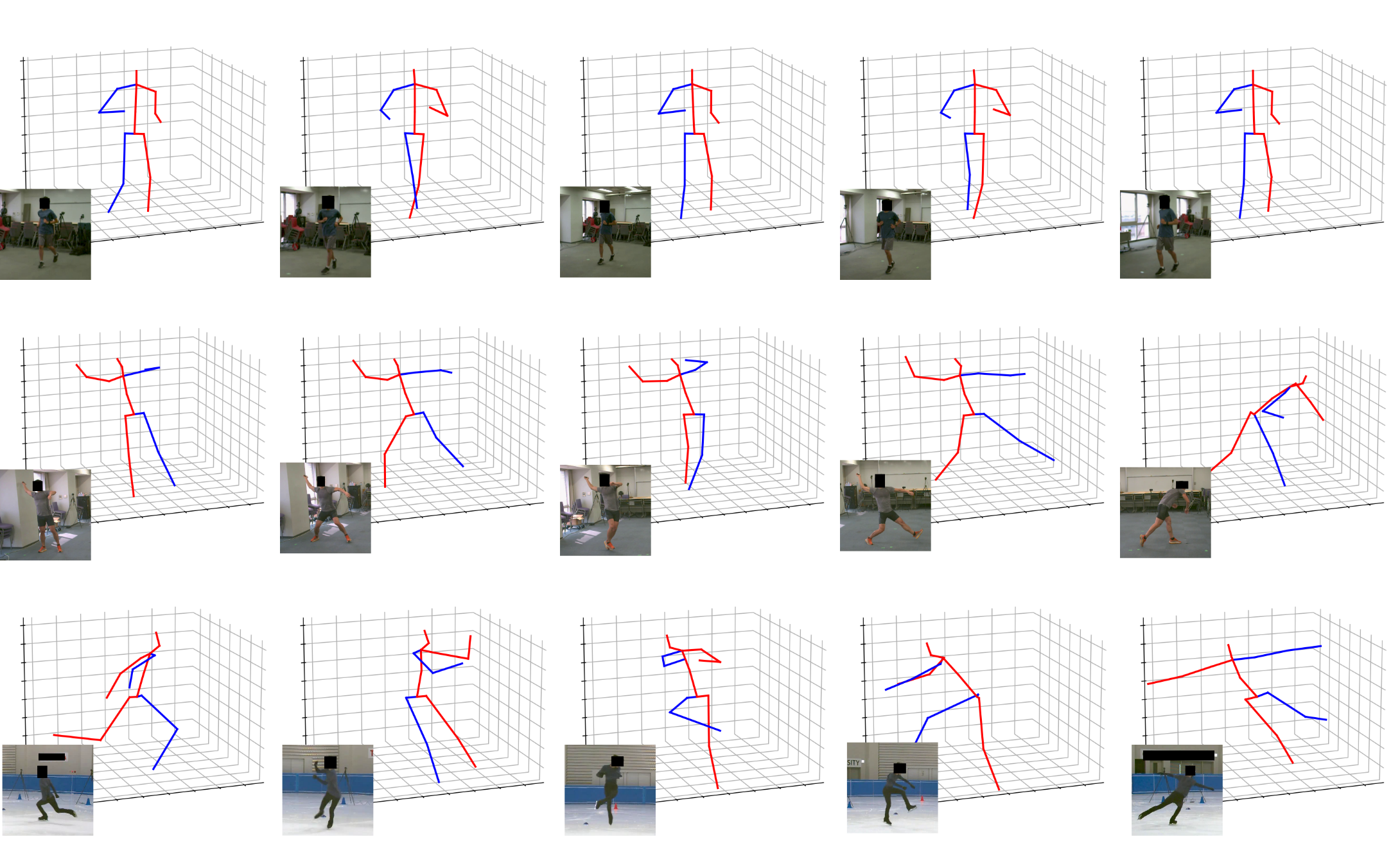}
    \caption{Examples of AthletePose3D data for running (top), track and field (middle), and figure skating (bottom). The poses are represented in centralized camera coordinates and rotated such that the motion progresses from left to right. The cropped image in the bottom left shows the posture as captured by the camera, with faces blacked out for privacy.}
\end{center}%
}]

\begin{abstract}
Human pose estimation is a critical task in computer vision and sports biomechanics, with applications spanning sports science, rehabilitation, and biomechanical research. While significant progress has been made in monocular 3D pose estimation, current datasets often fail to capture the complex, high-acceleration movements typical of competitive sports. In this work, we introduce \textbf{AthletePose3D}, a novel dataset designed to address this gap. AthletePose3D includes 12 types of sports motions across various disciplines, with approximately 1.3 million frames and 165 thousand individual postures, specifically capturing high-speed, high-acceleration athletic movements. We evaluate state-of-the-art (SOTA) monocular 2D and 3D pose estimation models on the dataset, revealing that models trained on conventional datasets perform poorly on athletic motions. However, fine-tuning these models on AthletePose3D notably reduces the SOTA model mean per joint position error (MPJPE) from 234mm to 98mm—a reduction of around 60\%. We also validate the kinematic accuracy of monocular pose estimations through waveform analysis, highlighting strong correlations in joint angle estimations but limitations in velocity estimation. Our work provides a comprehensive evaluation of monocular pose estimation models in the context of sports, contributing valuable insights for advancing monocular pose estimation techniques in high-performance sports environments.
\ifreview
  The dataset, code, and model checkpoints are available at: (to be updated post-acceptance)
\else
  The dataset, code, and model checkpoints are available at: \url{https://github.com/calvinyeungck/AthletePose3D}
\fi
\end{abstract}
\section{Introduction}
\label{sec:introduction}

Human pose estimation, a critical task in both computer vision \cite{suzuki2024pseudo,mehraban2024motionagformer,liu2025tcpformer,fang2024foul} and sports biomechanics \cite{haberkamp2022validity,fukushima2024potential,mirek2007assessment,sandbakk2012influence}, has seen significant advancements in recent years. Accurate 3D pose estimation plays a crucial role in understanding and analyzing human movement across various domains, including sports science \cite{haberkamp2022validity,fukushima2024potential}, rehabilitation \cite{hamilton2024comparison,menychtas2023gait}, and biomechanical research \cite{alvarez2022biometric}. Several 3D human pose datasets are available, such as Human3.6M \cite{h36m}, MPI-INF-3DHP \cite{3dhp}, and SportsPose \cite{SportsPose} (see Section \ref{ssec:3d_dataset} for a detailed overview).

Existing datasets \cite{h36m,3dhp,Humaneva,TotalCapture} provide valuable insights into human motion, consisting primarily of movements performed in controlled laboratory environments and daily activities. There are also sports-specific motion datasets \cite{SportsPose,ASPset}, which feature motions performed by amateurs. However, these datasets often fail to capture the intricate, high-acceleration movements that characterize competitive sports. Athletes engage in complex motions involving rapid changes in velocity, precise body positioning, and extraordinary biomechanical coordination—elements that are underrepresented in current datasets.

Additionally, capturing and analyzing sports movements presents significant challenges. Traditional motion capture systems, whether marker-based or markerless, are often expensive and time-consuming. While monocular pose estimation using deep learning models could offer a more effective solution, these models are typically trained and evaluated on datasets with limited movement complexity. Which could hinder their performance in the context of competitive sports. Furthermore, the kinematic estimates derived from state-of-the-art (SOTA) monocular 3D pose estimation models have not been sufficiently validated.

In response to these challenges, we introduce \textbf{AthletePose3D}, a sports motion dataset designed to bridge the gap in pose estimation data. This dataset includes 12 types of sports motions performed by athletes, comprising approximately 1.3 million frames and 165 thousand unique postures across disciplines such as running, track and field, and figure skating. Compared to conventional benchmarking datasets, AthletePose3D features notably higher speeds and accelerations, making it uniquely suited for athletic movement analysis.

Additionally, we provide a comprehensive evaluation of SOTA monocular 2D and 3D pose estimation models on AthletePose3D, demonstrating significant performance variations before and after fine-tuning on the dataset. We further validate the kinematic estimation capabilities of SOTA models through detailed waveform analysis. By establishing these benchmarks, this work aims to lay a solid foundation for advancing monocular pose estimation techniques in high-performance sports contexts.

\noindent Our key contributions are as follows:

\begin{itemize}

\item \textbf{Athlete Motion Dataset}: We introduce AthletePose3D, the first large-scale dataset capturing high-speed and high-acceleration 3D human poses across 12 unique sports actions performed by athletes.

\item \textbf{Comprehensive Model Evaluation}: We systematically benchmark state-of-the-art monocular 2D and 3D pose estimation models on our dataset, offering critical insights into their performance with athletic movements. Our experiments reveal that models trained exclusively on existing datasets perform poorly on athletic movements. However, after finetuning on AthletePose3D, error rates could be reduced by over 69\%, underscoring the need for sports-specific training data.

\item \textbf{Kinematic Validation}: We validate the estimated pose kinematics by comparing the kinematic waveforms against ground truth motion capture data, assessing the accuracy and reliability of monocular  pose estimation techniques in sports contexts. Our results show strong correlations in joint angle estimations, though some limitations remain in velocity estimation. 

\end{itemize}

The remainder of this paper is organized as follows.
Section \ref{sec:related_work} offers an overview of related works. The methods are detailed in Section \ref{sec:proposed_method}, followed by the details of AthletePose3D in Section \ref{sec:ap3d_dataset}. Section \ref{sec:experiments} presents the experimental results.
Finally, Section \ref{sec:conclusion} concludes the paper.

\section{Related work}
\label{sec:related_work}

\subsection{3D human pose datasets}
\label{ssec:3d_dataset}
\begin{table*}[ht]
    \centering
    \begin{tabular}{lcccccccccc}
        \hline
        Dataset & Environment & Subjects & Keypoints & Poses & Cameras & Markerless & Sync & FPS & Frames \\
        \hline
        Human3.6M \cite{h36m} & lab & 11 & 26 & 900K & 4 & $\times$ & hw & 50 & 3.6M \\
        MPI-INF-3DHP \cite{3dhp} & lab \& outdoor & 8 & 28 & 93K & 14 & $\checkmark$ & hw & 25/50 & 1.3M \\
        3DPW \cite{3dpw} & lab \& outdoor & 7 & 24 & 49K & 1 & $\times$ & sw & 30 & 51K \\
        HumanEva-I \cite{Humaneva} & lab & 6 & 15 & 78K & 7 & $\times$ & sw & 60 & 280K \\
        HumanEva-II \cite{Humaneva} & lab & 6 & 15 & 3K & 4 & $\times$ & hw & 60 & 10K \\
        TotalCapture \cite{TotalCapture} & lab & 5 & 25 & 179K & 8 & $\times$ & hw & 60 & 1.9M \\
        CMU Panoptic \cite{panoptic} & lab & 8 & 18 & 1.5M & 31 & $\checkmark$ & hw & 30 & 46.5M \\
        AIST++ \cite{AIST++} & lab & 30 & 17 & 1.1M & 9 & $\checkmark$ & sw & 60 & 10.1M \\
        ASPset-510 \cite{ASPset} & outdoor & 17 & 17 & 110K & 3 & $\checkmark$ & sw & 50 & 330K \\
        SportsPose \cite{SportsPose} & lab \& outdoor & 24 & 17 & 177K & 7 & $\checkmark$ & hw & 90 & 1.5M \\
        AthletePose3D (ours) & lab \& ice rink & 8 & 55/86 & 165K & 4/8/12 & $\checkmark$ & hw & 60/120 & 1.3M \\
        \hline
    \end{tabular}
    \caption{Comparison of the AthletePose3D dataset with other 3D pose datasets for monocular 3D pose estimation. The configuration of AthletePose3D (Keypoints, Cameras, FPS) was adapted to suit the specific environment and sports captured.}
    \label{tab:datasets}
\end{table*}

Large-scale 3D human pose datasets have been fundamental for training and evaluating monocular 3D pose estimation models. Table \ref{tab:datasets} provides a detailed summary of existing 3D human pose datasets. Significant efforts have been made to develop large-scale datasets, with Human3.6M \cite{h36m} and MPI-INF-3DHP \cite{3dhp} being among the most well-known benchmarks for monocular 3D pose estimation. Broadly, 3D human pose datasets are collected using two mainstream methods: marker-based motion capture, which employs optical tracking systems and infrared cameras (e.g., Human3.6M \cite{h36m}, 3DWP \cite{3dpw}, HumanEva \cite{Humaneva}, TotalCapture \cite{TotalCapture}), and markerless motion capture, which relies on industrial-grade cameras and computer vision algorithms (e.g., MPI-INF-3DHP \cite{3dhp}, CMU Panoptic \cite{panoptic}, SportsPose \cite{SportsPose}, FS-Jump3D \cite{tanaka20243d}, WorldPose \cite{jiang2024worldpose}). 

Additionally, alternative approaches exist, including inferring 3D poses from 2D annotations (e.g., 3DSP \cite{AutoSoccerPose}), estimating 3D poses via triangulation from multiple 2D pose detections (e.g., ASPset-510 \cite{ASPset}), and reconstructing 3D poses from real-world 3D human pose data (e.g., AIST++ \cite{AIST++}). Whilst various methods exist for capturing 3D human posture, marker-based and markerless approaches remain the gold standard and are widely used in sports science \cite{takigawa2023factors,makino2022kinematic} and biomechanics \cite{mirek2007assessment,sandbakk2012influence} studies.

For 3D human posture in sports, SportsPose \cite{SportsPose} and ASPset-510 \cite{ASPset} are currently the largest datasets designed to facilitate monocular 3D human posture estimation. Other datasets, such as 3DSP \cite{AutoSoccerPose} and WorldPose \cite{jiang2024worldpose}, provide posture data from professional soccer matches for posture analysis and pose tracking, respectively, while FS-Jump3D \cite{tanaka20243d} focuses on enhancing figure skating 3D pose estimation. 

However, existing datasets are often not specifically designed for monocular 3D pose estimation or lack high-speed, high-acceleration movements that are essential in competitive sports. To address these limitations, we introduce AthletePose3D, a comprehensive dataset encompassing 12 types of sports actions performed by athletes ranging from amateur athletes (athletes without remuneration) to professional athletes. See Section \ref{sec:ap3d_dataset} for more details.

\subsection{Monocular 2D \& 3D human pose estimations}
\label{ssec:pose_estimation_review}
In computer vision, both monocular 2D \& 3D human pose estimations are a well-established research topic. 2D human pose estimation approaches are generally categorized into bottom-up and top-down approaches. Bottom-up methods \cite{cao2017realtime,cheng2020higherhrnet,geng2021bottom} first detect keypoints in an image and then group them to form 2D poses, whereas top-down methods \cite{jiang2023rtmpose,sun2019deep,yang2023effective} begin by detecting individuals using object detection models before estimating poses within cropped bounding boxes. Due to their effectiveness, top-down approaches are often preferred. 

In 2D pose estimation models, early methods were primarily CNN-based, proposing architectures such as ResNet \cite{xiao2018simple}, HRNet \cite{sun2019deep}, and SwinPose \cite{liu2021swin}. Later approaches introduced transformers \cite{vaswani2017attention} to refine feature decoding from CNN backbones, including cascade transformers \cite{li2021pose}, TransPose \cite{yang2021transpose}, TokenPose \cite{li2021tokenpose}, and UniFormer \cite{iclr2022UniFormer}. Further advancements replaced CNNs entirely with transformer-based architectures, as seen in HRFormer \cite{yuan2021hrformer} and ViTPose \cite{xu2022vitpose}, which employ vision transformers \cite{dosovitskiy2020image} for pose estimation. More recently, the resurgence of CNN models \cite{liu2022convnet,liu2022more,rao2022hornet,yang2022focal}, has led to a renewed focus on CNN-based methods like RTMPose \cite{jiang2023rtmpose} and MogaNet \cite{iclr2024MogaNet}. 

Monocular 3D pose estimation methods can be broadly classified into direct 3D pose estimation \cite{chun2023learnable, iskakov2019learnable, reddy2021tessetrack} and 2D-to-3D lifting \cite{mehraban2024motionagformer, yang2023effective, sun2019deep}, where the latter involves reconstructing 3D poses from 2D keypoint coordinates. Early approaches to monocular 3D pose estimation were based on Hidden Markov Models \cite{lehrmann2014efficient, trabelsi2013unsupervised} and graphical models \cite{kruger2016efficient, song2001unsupervised}. With the rise of deep learning, Fully Connected Networks \cite{ci2023gfpose, martinez2017simple} were introduced, followed by temporal-based \cite{cheng20203d, pavllo20193d} and graph-based \cite{cai2019exploiting, ci2019optimizing, wang2020motion} convolutional networks, which improved the modeling of sequential dependencies and inter-keypoint relationships.

More recently, transformer-based methods have become the dominant approach, achieving state-of-the-art performance. PoseFormer \cite{zheng20213d} pioneered this direction, leading to further advancements with models such as \cite{zhao2023poseformerv2,li2022mhformer,zhang2022mixste,tang20233d,shan2023diffusion,shan2022p,einfalt2023uplift,li2022exploiting,tang20233d,qian2023hstformer}. The latest models, KTPFormer \cite{peng2024ktpformer} integrates kinematics and trajectory priors to improve motion representation, MotionAGFormer \cite{mehraban2024motionagformer} introduces a hybrid approach, combining self-attention with graph convolution to better capture both spatial and temporal dependencies, and TCPFormer \cite{liu2025tcpformer} further enhances temporal modeling by leveraging pose proxies to refine sequential feature learning. Additionally, self-supervised learning has been extensively explored in monocular 3D pose estimation \cite{li20213d, wang2022contrast, endo2022gaitforemer, nekoui2021enhancing, yang2020transmomo, zhu2022mocanet}.

Although many monocular pose estimation models have been proposed, they are often benchmarked on datasets featuring everyday movements. High-speed sports movements, however, present unique challenges, and the performance of these models on such movements remains unclear. Therefore, in this study, we select the best-performing monocular 2D \& 3D models from corresponding benchmark datasets and fine-tune them on the AthletePose3D dataset to evaluate their performance. 


\subsection{Validate kinematic from pose estimation}
\label{ssec:valide_kinematic_review}
Kinematic analysis has been widely used in various fields, including sports analysis \cite{haberkamp2022validity,fukushima2024potential}, rehabilitation \cite{hamilton2024comparison,menychtas2023gait}, and biometric recognition \cite{alvarez2022biometric}. As a result, numerous studies \cite{menychtas2023gait,washabaugh2022comparing,hamilton2024comparison,ceriola2023comparison} have aimed to validate the use of 2D and 3D motion estimation for kinematic analysis.

In sports research, \cite{fukushima2024potential} compares athletic and sports motion kinematics using a marker-based motion capture system and 3D pose estimation derived from triangulated OpenPose \cite{8765346openpose} 2D pose estimation across multiple cameras. Similarly, \cite{haberkamp2022validity} investigates single-leg squat motions by comparing the waveforms of movements captured via OpenPose \cite{8765346openpose} 2D pose estimation and 3D motion capture.

Although models like OpenPose \cite{8765346openpose}, Mediapipe \cite{lugaresi2019mediapipe}, and HRNet \cite{sun2019deep} are frequently used in sports science and kinematic validation studies, they do not represent the state-of-the-art in 2D pose estimation. Moreover, monocular 3D pose estimation models have received limited attention in validation studies. Therefore, this study evaluates both state-of-the-art monocular 2D and 3D pose estimation models for sports motion kinematic waveform using the newly introduced AthletePose3D dataset.

\section{Methodology}
\label{sec:proposed_method}

\subsection{Motion capture setup}
For data capture, we employed a multi-camera system consisting of 4, 8, and 12 high-speed cameras (Miqus Video, Qualisys Inc.) for running, track and field, and figure skating, respectively. The number of cameras was selected based on the complexity of the movements in each sport, ensuring sufficient coverage to accurately track the athletes' motion from multiple angles. The camera layout was strategically designed to minimize occlusions and accommodate the natural movement patterns of athletes without imposing any restrictions on their performance.

All cameras were hardware-synchronized to ensure precise temporal alignment of captured frames, which is crucial for accurate 3D motion reconstruction. The calibration process was conducted using Qualisys Track Manager (Qualisys), employing a wand-type calibration kit to establish accurate spatial relationships between the cameras. This setup achieved a calibration error of less than 1 mm, ensuring high precision in 3D motion extraction.

The frame rates were chosen based on the movement characteristics of each sport. Running, which involves rapid limb movements and high-speed motion, was recorded at 120 FPS to capture fine details. Track and field events, which typically involve a mix of explosive and sustained movements, were recorded at 60 FPS to balance temporal resolution with data storage efficiency. Similarly, figure skating, which features both dynamic jumps and continuous fluid motion, was also recorded at 60 FPS, as it provides sufficient detail while maintaining manageable data volumes. The captured videos had a primary resolution of 1920 × 1080, ensuring high-quality motion capture and enabling precise monocular 3D pose estimation.

\begin{table*}[!ht]
\centering
\begin{tabular}{lcccccccccc}
\hline
Sport            & Env.  &  Act. Types & Subs  & Kpts & Cams & Poses &FPS & Frames & Comp. Level  & Representative   \\ \hline
Running          & Lab   & 1  & 3 & 55   & 4    & 40K & 120   & 161K   & Inter-university   & University  \\ 
Track \& Field   & Lab   & 5  & 1 & 86   & 8    & 52K & 60   & 416K   & International      & National   \\ 
Figure Skating   & Ice rink   & 6 & 4  & 86   & 12   & 73K & 60   & 700K   & national   & University  \\ \hline
\end{tabular}
\caption{Details of the 3D pose for various sports in AthletePose3D, including subjects' competition experience and team representation. }
\label{tab:sports_datasets}
\end{table*}

\subsection{Benchmarking monocular 2D \& 3D pose estimation models}

To benchmark both monocular 2D and 3D pose estimation models, we considered the best-performing models on representative datasets, taking into account their architectures. These models were fine-tuned and evaluated using the AthletePose3D dataset, which was split into training, validation, and test sets with a 60/20/20 ratio. The training set was used for model training, and the validation set results were reported.

For \textbf{2D pose estimation}, we selected HRNet~\cite{sun2019deep} and SwinPose~\cite{liu2021swin} as representatives of conventional CNN-based methods. ViTPose~\cite{xu2022vitpose} and UniFormer~\cite{iclr2022UniFormer} were chosen to represent transformer-based approaches, while MogaNet~\cite{iclr2024MogaNet} was included as a recent CNN-based resurgence. All selected models were state-of-the-art (SOTA) on either the COCO~\cite{cocodataset} or OCHuman~\cite{zhang2019pose2seg} datasets.  

For finetuning COCO dataset pretrained 2D pose estimation model with AthletePose3D, we projected the 3D keypoints into 2D keypoints in the COCO format, omitting facial keypoints (kpt 1 to 4) and setting their loss weights to zero during training. The fine-tuning process followed \cite{sun2019deep}, with training conducted for 20 epochs. The learning rate was initially set to $1 \times 10^{-4}$, reduced to $1 \times 10^{-5}$ after 10 epochs, and further decreased to $1 \times 10^{-6}$ after 15 epochs. All models used an input resolution of $384\times288$, except for ViTPose, where the $384\times288$ variant was unavailable, and thus, the $256\times192$ resolution was used. Only the model variants with the highest reported Average Precision (AP) were selected.  

The performance is assessed using the Percent of Detected Joints (PDJ) evaluation metric \cite{toshev2014deeppose}. A keypoint is considered detected if the normalized distance between the predicted and ground truth keypoints is below a threshold. Following \cite{tompson2014joint,AutoSoccerPose}, the normalization factor is defined as the distance between the center of the shoulders and the center of the hips. A threshold of 0.2 is selected as a cutoff, as higher thresholds tend to yield overly high PDJ values. Additionally, the area under the PDJ curve (AUC) is computed to evaluate performance across varying thresholds.

For Monocular \textbf{3D pose estimation}, we select MotionAGFormer \cite{mehraban2024motionagformer} and TCPFormer \cite{liu2025tcpformer}, as most monocular 3D pose estimation models are transformer-based. The selected models are SOTA on the Human3.6M \cite{h36m} and MPI-INF-3DHP \cite{3dhp} datasets.  

For fine-tuning, the 3D keypoints follow the Human3.6M format, and preprocessing is performed according to \cite{ci2020locally,motionbert2022,mehraban2024motionagformer,liu2025tcpformer}. Specifically, the 3D poses in world coordinates are converted to camera coordinates using the extrinsic parameters. The $(X, Y)$ coordinates are then projected onto the image plane using the intrinsic parameters. The $Z$ coordinate in image space is obtained by scaling the $Z$ coordinate in camera space. The scaling factor is determined by minimizing the distance between the scaled $(X, Y)$ camera coordinates and their corresponding image coordinates. 

We compare three training strategies: training on Human3.6M, training on AthletePose3D, and training on Human3.6M followed by fine-tuning on AthletePose3D. Given the dataset size, both training and fine-tuning are conducted for 60 epochs. Furthermore, considering the typical video length of sports motion sequences (see Section \ref{ssec:compare_exist_dataset}), we use the model variant that processes 81 frames.  

Performance is assessed using the Mean Per Joint Position Error (MPJPE), which measures the per-joint distance between the estimated 3D poses and the ground truth, after aligning both poses by setting the hip joint center as the origin. Additionally, the Procrustes-aligned MPJPE (P-MPJPE) applies a rigid transformation to the predicted pose before computing the error.

\subsection{Estimated pose kinematic validation}
In general, monocular pose estimation provides a more efficient alternative to conventional motion capture methods. Therefore,
the best-performing monocular 2D and 3D pose estimation models on the AthletePose3D validation set (see Sections \ref{ssec:2d_result} and \ref{ssec:3d_result}) were selected for evaluation. Using these models, 2D and 3D poses were estimated from clips in the AthletePose3D test set. Next, limb joint velocity and joint angles were computed, as these joints are central to kinematic analysis and tend to have the highest errors. To smooth the kinematic data, a fourth-order Butterworth low-pass digital filter with zero-phase shift at 8 Hz was applied, following \cite{takigawa2023factors,makino2022kinematic}.

Since coordination differences affect absolute kinematic values, the analysis focused on the waveform of kinematic data (e.g., the temporal series of wrist velocity in a javelin throw) rather than exact values, as the waveform can reveal valuable insights. To validate the motion waveform, paired t-test statistical parametric mapping (SPM) with a significance level of \(\alpha = 0.05\) was performed, and Pearson correlation was reported for a more intuitive interpretation.

\section{AthletePose3D dataset}
\label{sec:ap3d_dataset}

\subsection{Dataset statistics}

The AthletePose3D dataset aims to capture high-speed, high-acceleration, and professional sports motions, with a detailed summary provided in Table \ref{tab:sports_datasets}. The collected motions are categorized into three sports—running, track and field, and figure skating—comprising 12 unique motion types. Specifically, track and field includes shot put, glide shot put, javelin throw, discus, and spin discus, while figure skating consists of Axel, Salchow, Toe Loop, Loop, Flip, and Lutz. Running and track and field are among the most widely studied sports in sports science, whereas figure skating involves highly dynamic and complex movements, making it particularly challenging for 3D motion estimation and a frequent benchmark in recent models.

A total of eight athletes participated in data collection: three for running, one for track and field, and four for figure skating. Their expertise ranged from university-level representatives to national-level representatives, with experience spanning inter-university and international competitions.
The University's ethics committee approved this data collection, and written informed consent was obtained from all participants.
Building on previous measurements of running
\ifreview
  [anon. ref.]
\else
  \cite{suzuki2024pseudo} 
\fi
and figure skating 
\ifreview
  [anon. ref.], 
\else
  \cite{tanaka20243d}, 
\fi
we present previously unpublished data, including camera angles and monocular 3D pose estimation required information (e.g., camera parameters and valid frames), which were collected in our earlier research.
In total, approximately 1.3 million frames of data were collected, capturing around 165K individual postures, including 40K from running, 52K from track and field, and 73K from figure skating. 

\begin{figure}[t]
  \centering
   \includegraphics[width=1\linewidth]{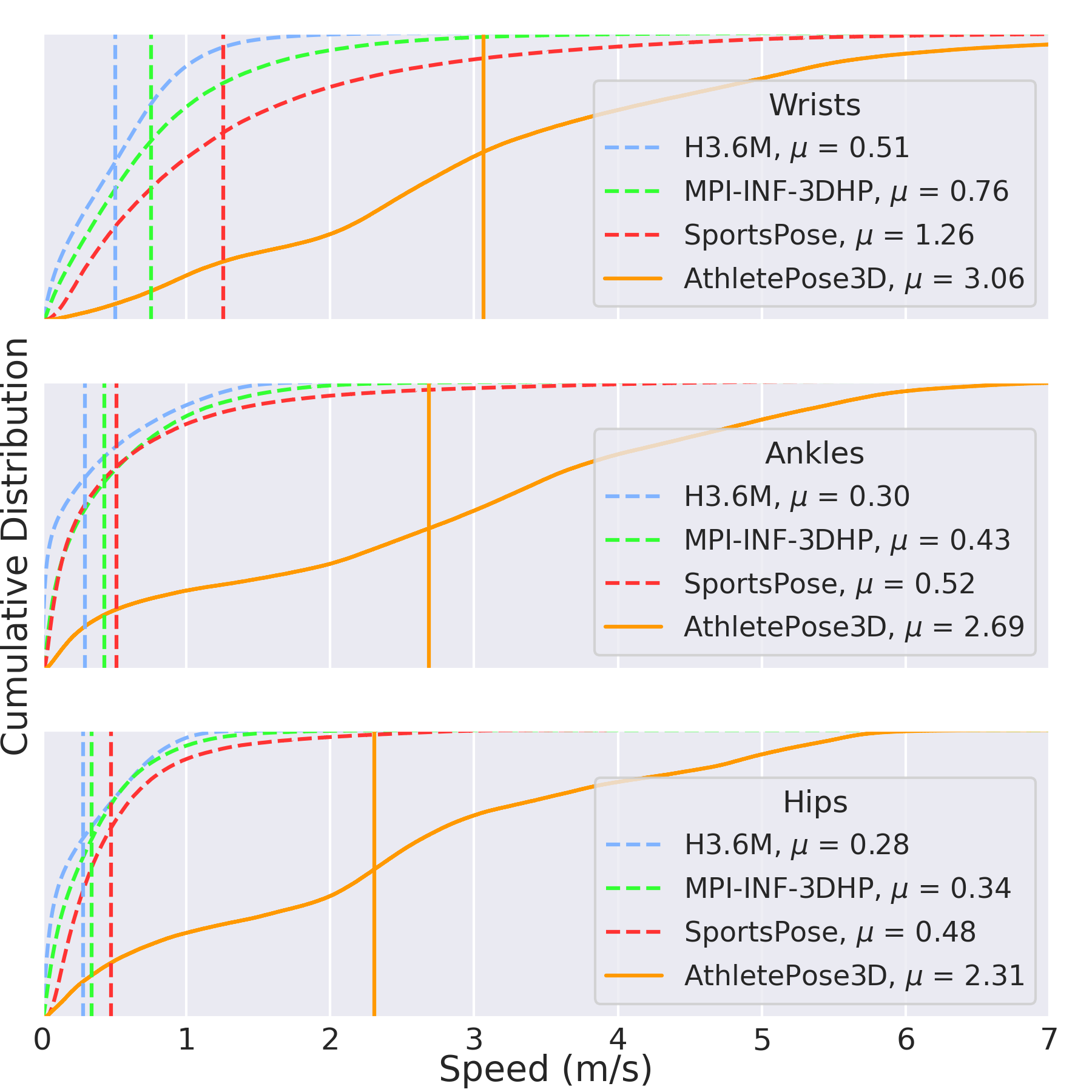}
   \caption{Comparison of the cumulative distribution function (CDF) of wrist, ankle, and hip speeds across 3D pose datasets. A lower CDF curve indicates a higher speed relative to the other curves, with the mean speed ($\mu$) marked by a vertical line.}
   \label{fig:speed_cdf}
\end{figure}

\begin{figure}[t]
  \centering
   \includegraphics[width=1\linewidth]{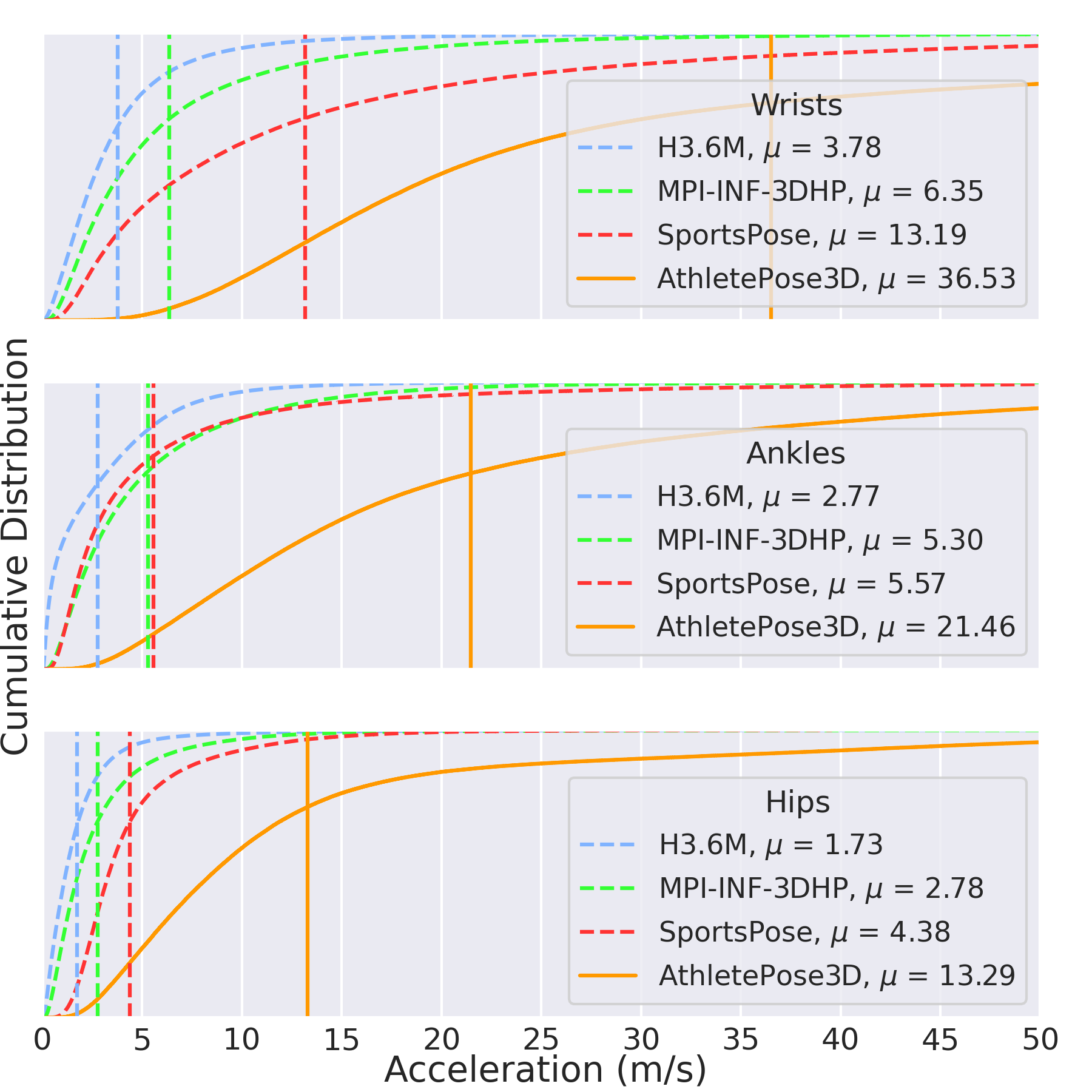}

   \caption{Comparison of the cumulative distribution function (CDF) of wrist, ankle, and hip acceleration across 3D pose datasets. A lower CDF curve indicates a higher acceleration relative to the other curves, with the mean speed ($\mu$) marked by a vertical line.}
   \label{fig:acc_cdf}
\end{figure}

\begin{figure}[t]
  \centering
   \includegraphics[width=1.1\linewidth]{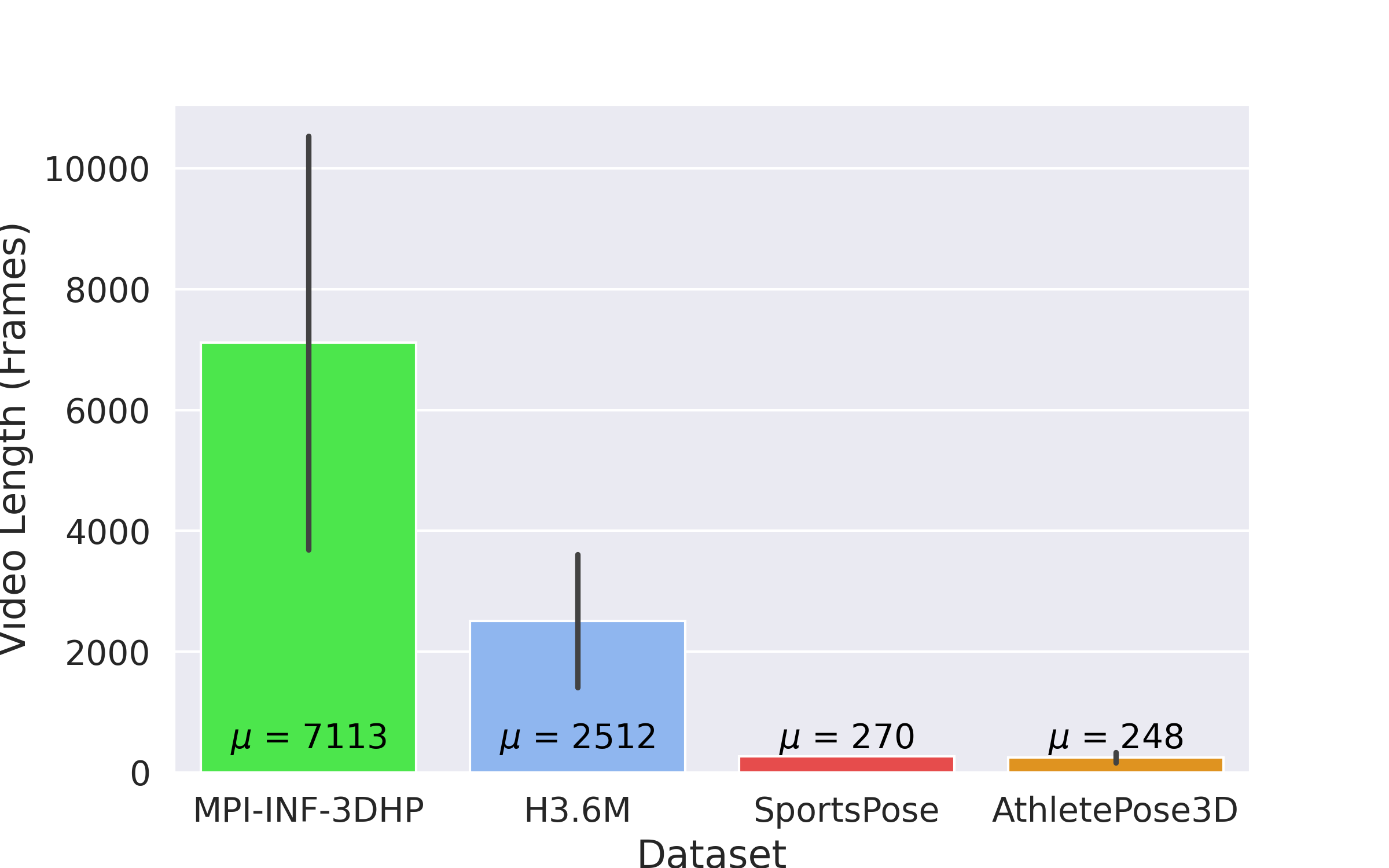}

   \caption{Video lengths of 3D pose datasets. The black bar represents the standard deviation, calculated using the Seaborn Python package.}
   \label{fig:video_length}
\end{figure}

\subsection{Compare with existing dataset}
\label{ssec:compare_exist_dataset}
When comparing AthletePose3D with existing 3D human pose datasets such as Human3.6M \cite{h36m}, MPI-INF-3DHP \cite{3dhp}, and SportsPose \cite{SportsPose}, joint speed, joint acceleration, and video length are key considerations. Figures \ref{fig:speed_cdf} and \ref{fig:acc_cdf} show the cumulative distribution of joint speed and acceleration for the wrists, ankles, and hips, respectively. It can be observed that both the distribution and mean values in AthletePose3D are significantly higher than those in previous datasets, including SportsPose. This highlights the difference in motion speed and acceleration between everyday movements and sports motions performed by athletes in competitive-like settings, emphasizing the need for a dataset like AthletePose3D.

Another key distinction between 3D human pose datasets, often overlooked, is the video length, as shown in Figure \ref{fig:video_length}. Datasets such as Human3.6M and MPI-INF-3DHP, which focus on everyday movements, typically feature videos that extend beyond a thousand frames. In contrast, datasets capturing sports motions, like SportPose and AthletePose3D, generally consist of shorter videos, averaging around 250 frames. This reflects the dynamic nature of sports movements, which occur within seconds and are designed to capture specific actions of interest. Given this characteristic, we selected a shorter input size for the 3D pose estimation models utilized in this study, 81 frames.



\section{Experiments results}
\label{sec:experiments}

\subsection{2D pose estimation models performance}
\label{ssec:2d_result}

\begin{table*}[]
\centering
\begin{tabular}{lccccccccc}
\hline
Model     & Shoulder & Elbow & Wrist & Hip  & Knee & Ankle & PDJ$_{[.1]}$    & PDJ-AUC$_{[.0:.2]}$  & AP$_{[.5:.95]}$  \\\hline
HRNet (CVPR'2019) \cite{sun2019deep}     & 88.3 & 77.9
  & 86.7 & 90.1 & 93.4 & 96.0  & 88.7 & 74.8 & 94.0 \\
SwinPose (ICCV’2021) \cite{liu2021swin}  & 93.9     & 84.2  & 86.3  & 95.0 & 89.4 & 95.5  & 90.7 & 76.2 & 95.2 \\
VitPose (NeurIPS'2022) \cite{xu2022vitpose}  & 97.4     & 88.7  & 91.0  & 98.1 & 96.6 & 98.3  & 95.0 & 81.4 & \textbf{95.8} \\
UniFormer (ICLR'2022) \cite{iclr2022UniFormer} & 97.2     & 89.4  & 91.2  & 98.1 & 97.2 & 98.0  & 95.2 & 80.9 & 94.9 \\
MogaNet (ICLR'2024) \cite{iclr2024MogaNet}   & \textbf{97.7}     & \textbf{90.6}  & \textbf{92.0}  & \textbf{98.2} & \textbf{97.4} & \textbf{98.4}  & \textbf{95.7} & \textbf{81.7} & 95.6\\
\hline
\end{tabular}
\caption{2D pose estimation performance on AthletePose3D validation set, with the top-performing result highlighted in bold. Ranked by PDJ in descending order. Each column for body parts indicates the average PDJ for the specified body parts, while the PDJ column denotes the mean PDJ across all 12 keypoints (keypoints 5-16 in COCO format). The subscript denotes the threshold used for the metrics.}
\label{tab:2d_result}
\end{table*}

\begin{figure}[t]
  \centering
   \includegraphics[width=1.1\linewidth]{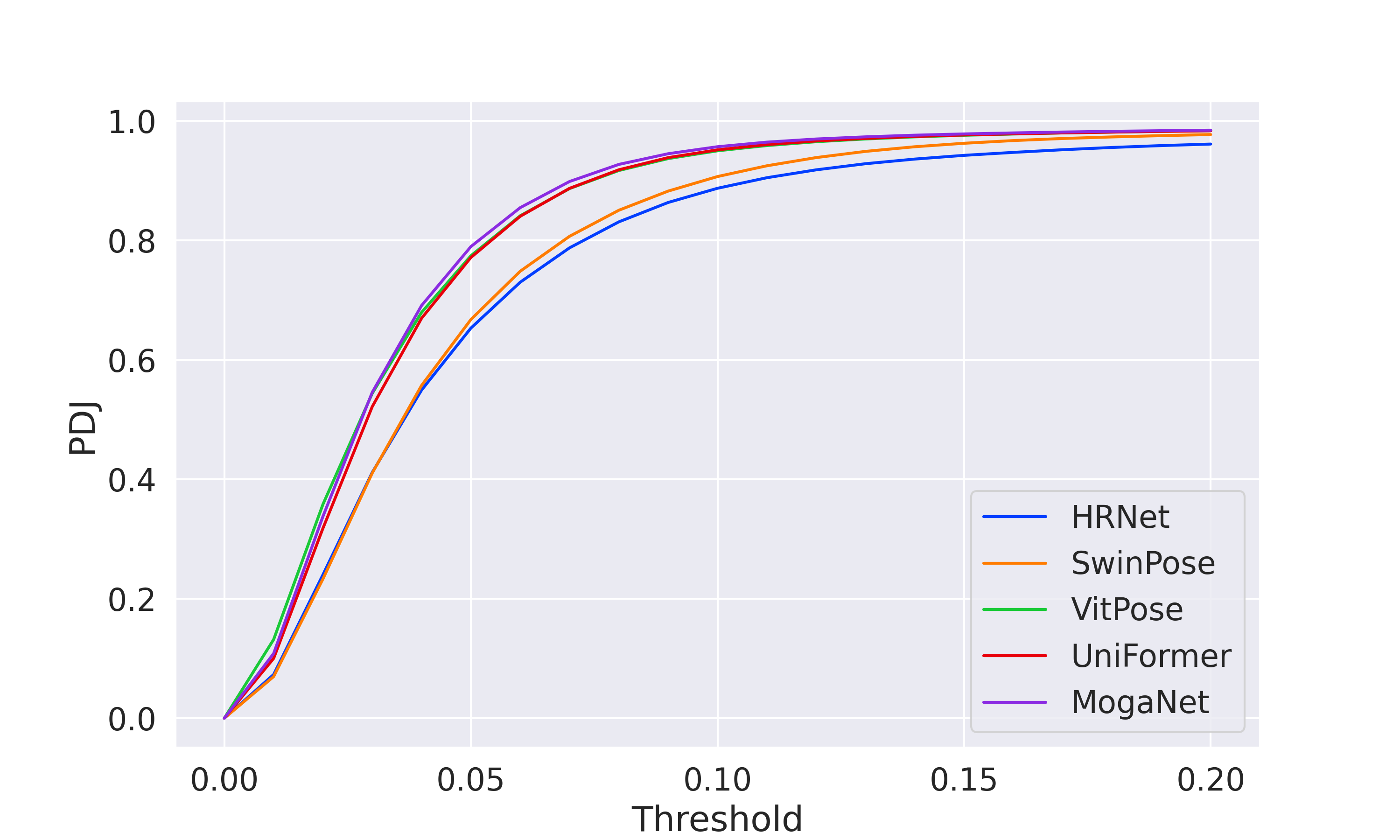}

   \caption{PDJ curve for 2D pose estimation model.}
   \label{fig:pdj_auc}
\end{figure}

\begin{table*}[]
\centering
\begin{tabular}{lccccccccc}
\hline
Model          & Dataset             & Shoulder       & Elbow          & Wrist          & Hip            & Knee           & Ankle           & MPJPE          & P-MPJPE       \\ \hline
MotionAGFormer & H3.6M               & 250.09
         & 234.27
         & 254.57
         & 74.87
          & 320.39
         & 560.23
          & 257.26
         & 95.67
         \\
(WACV’2024) \cite{mehraban2024motionagformer}               & AP3D       & 118.58
          & 122.45
          & 88.48
          & 30.45
          & 112.84
          & 168.10
          & 100.12
          & 31.84
         \\
               & H3.6M+AP3D & 117.59
          & 121.82
          & 86.83
          & 29.63
 & \textbf{110.08}
          & 165.16
          & 98.62
          & 30.54
          \\
TCPFormer      & H3.6M               & 260.41
         & 257.32
         & 299.24
         & 80.47
          & 253.15
         & 324.96
          & 234.20
         & 101.80
         \\
(AAAI’2025) \cite{liu2025tcpformer}                & AP3D       & \textbf{117.16}
          & 121.25
 & 87.56
          & 30.24
          & 110.40
          & \textbf{165.15}
          & 98.77
          & 30.67
          \\
               & H3.6M+AP3D & 117.24
 & \textbf{120.87}
          & \textbf{86.08
} & \textbf{29.39}
          & 110.13
 & 164.29
 & \textbf{98.26
} & \textbf{29.91
}\\
\hline
\end{tabular}
\caption{Monocular 3D pose estimation (with 2D pose ground truth) performance on AthletePose3D validation set, with the top-performing result highlighted in bold. Each column for body parts indicates the average JPE for the specified body parts, while the MPJPE and P-MPJPE are also known as Protocols 1 and 2, respectively.}
\label{tab:3d_result}
\end{table*}

The performance of various 2D pose estimation models on the AthletePose3D validation set was summarized in Table \ref{tab:2d_result}, ranked by their average PDJ across body parts. MogaNet \cite{iclr2024MogaNet} achieved the highest performance, with the best PDJ values for each joint and an overall PDJ of 95.7. UniFormer \cite{iclr2022UniFormer} followed closely, trailing slightly behind MogaNet in PDJ. ViTPose \cite{xu2022vitpose} and SwinPose \cite{liu2021swin} delivered competitive results, with ViTPose leading in AP. HRNet \cite{sun2019deep} lagged behind, particularly in Elbow predictions.

The PDJ curves in Figure \ref{fig:pdj_auc} provide a more detailed comparison of model performance across different thresholds. As expected, increasing the threshold resulted in improved PDJ scores for all models, as larger thresholds allowed for greater error tolerance. MogaNet and UniFormer maintained strong performance throughout, consistently yielding the highest PDJ scores across all thresholds. ViTPose and SwinPose performed well but showed slightly lower results, with HRNet trailing behind in terms of joint position accuracy, especially at smaller thresholds.

The results highlighted that CNN-based models like MogaNet and transformer-based models like UniFormer performed similarly well, suggesting that both architectural approaches could achieve high accuracy in 2D pose estimation tasks. This finding indicated that the choice between convolutional and transformer-based architectures could depend on factors such as computational efficiency and generalization to unseen poses, rather than absolute performance.

\subsection{Monocular 3D pose estimation models performance}
\label{ssec:3d_result}

The results in Table \ref{tab:3d_result} demonstrated the impact of different training datasets on monocular 3D pose estimation performance when evaluated on the AthletePose3D (AP3D) validation set. Models trained solely on Human3.6M (H3.6M) performed significantly worse than those trained on AP3D or a combination of both datasets. MotionAGFormer, when trained on H3.6M, exhibited high joint position errors, particularly for the Knee (320.39 mm) and ankle (560.23 mm), resulting in an MPJPE of 257.26 mm. Similarly, TCPFormer trained on H3.6M struggled with high errors, especially in the Ankle  (324.96 mm), although it achieved a slightly lower MPJPE of 234.20 mm.

Training on AP3D substantially improved performance for both models, reducing MPJPE by around 60\% compared to training solely on H3.6M. TCPFormer achieved an MPJPE of 98.77 mm and a P-MPJPE of 30.67 mm, outperforming MotionAGFormer (100.12 mm MPJPE, 31.84 mm P-MPJPE). The most notable improvements were observed in lower-body joints, particularly the knee and ankle, where errors dropped from over 300 mm (H3.6M-trained models) to approximately 110 mm and 170 mm, respectively.

Utilizing both H3.6M and AP3D for training led to further refinements, with TCPFormer achieving the lowest overall errors (MPJPE: 98.26 mm, P-MPJPE: 29.91 mm). The best per-joint accuracy was also observed in this setting, with notable improvements in the wrist (86.08 mm), knee (110.13 mm), and ankle (164.29 mm). These results highlighted that training on sports-specific datasets like AP3D was crucial for achieving accurate monocular pose estimation in athletic movements, and combining diverse datasets further enhanced robustness.

\subsection{Estimated pose kinematic validation}

\begin{table}[]
\centering
\begin{tabular}{lcccc}
\hline 
                       & \multicolumn{2}{l}{\begin{tabular}[c]{@{}c@{}}2D Pose Estimation \\ vs \\ GT Motion Capture\end{tabular}} & \multicolumn{2}{l}{\begin{tabular}[c]{@{}c@{}}3D Pose Estimation \\ vs \\ GT Motion Capture\end{tabular}} \\
                      & $r$                                            & p-value                                                                              & $r$                                                  & p-value                                           \\
\hline
\textbf{Velocity}                                                            &                                                   &                                                       &                                                       &                                                   \\
Upper limbs                                            & 0.47                                          & 0.032                                                 & 0.28                                     & 0.036                                             \\
Lower limbs                                            & 0.45                                         & 0.027                                                 & 0.11                                               & 0.004                                             \\
\textbf{Joint Angle}                                          &                                                         &                                                                                    &                                                   \\
Upper limbs                                     & 0.75                                           & 0.021                                        & 0.90                                              & 0.001                                    \\
Lower limbs                                               & 0.66                                           & $<$0.001                                                & 0.82                                              & 0.044         \\
\hline
\end{tabular}
\caption{Limbs kinematic validation result from monocular pose estimation models. To perform the paired t-test SPM, the kinematic temporal sequences were interpolated linearly into the same length.}
\label{tab:kinematic_validation}
\end{table}

Table \ref{tab:kinematic_validation} presents the results of the paired t-test SPM analysis and Pearson correlation ($r$), comparing the estimated kinematic waveforms from monocular 2D and 3D pose estimation model, MogaNet \cite{iclr2024MogaNet} and TCPformer \cite{liu2025tcpformer}, to ground-truth (GT) motion capture data. If the p-value is lower than the significance level (i.e., $p < 0.05$), it suggests a significant difference between the pose estimation waveforms and the GT motion capture waveform. Pearson correlation provides an additional measure of similarity, where higher values indicate a stronger relationship between estimated and GT waveforms.  

For \textbf{velocity estimation}, the 2D pose model showed moderate correlation with GT data for both the upper (\(r = 0.47\), \(p = 0.032\)) and lower limbs (\(r = 0.45\), \(p = 0.027\)), despite significant differences detected by the paired t-test SPM. Similarly, the 3D pose model exhibited a lower correlation for upper limb velocity (\(r = 0.28\), \(p = 0.036\)) and a weak correlation for lower limb velocity (\(r = 0.11\), \(p = 0.004\)), with significant waveform differences. These results suggested that while the estimated velocity waveforms captured some temporal trends, they still significantly deviated from GT motion capture.

For \textbf{joint angles}, the correlations were substantially higher, particularly for the 3D pose model. The 2D model achieved strong correlations for upper limb (\(r = 0.75\), \(p = 0.021\)) and lower limb (\(r = 0.66\), \(p < 0.001\)) joint angles. The 3D model further improved upon this, showing very strong correlations for upper limb (\(r = 0.90\), \(p = 0.001\)) and lower limb (\(r = 0.82\), \(p = 0.044\)) joint angles. However, despite these high correlations, all joint angle waveforms still exhibited statistically significant differences from GT data according to the paired t-test SPM, indicating residual discrepancies in waveform shape.

Since all paired t-test results yielded \(p < 0.05\), the null hypothesis of no difference was rejected for all comparisons, meaning that both monocular 2D and 3D pose estimation models produced waveforms that significantly differed from GT motion capture. However, high correlation values, particularly for joint angles, suggested that these models still captured meaningful kinematic trends. These results highlighted the potential of monocular pose estimation as an efficient alternative to conventional motion capture, though further refinements were necessary, especially for velocity estimation.





\section{Conclusion}
\label{sec:conclusion}
A key challenge in human pose estimation for competitive sports is the lack of datasets capturing high-speed, high-acceleration movements, as well as the insufficient validation of kinematic estimates from SOTA models. Existing datasets often fail to represent the complexities of athletic actions, leading to unreliable monocular pose estimations.

To address this, we introduced AthletePose3D, a large-scale dataset featuring 12 sports actions performed by athletes, with over 1.3 million frames and 165 thousand individual postures. Our evaluation of SOTA monocular 2D and 3D pose estimation model, MogaNet \cite{iclr2024MogaNet} and TCPformer \cite{liu2025tcpformer}, revealed poor performance on athletic movements, but fine-tuning on AthletePose3D improved accuracy, reducing error rates by over 70\%. Kinematic validation showed strong joint angle correlations, though velocity estimation remains a challenge.

AthletePose3D provides a crucial resource for advancing monocular pose estimation in sports, offering a foundation for more accurate motion analysis in sports science, biomechanics, and rehabilitation. This can lead to improved performance optimization and injury prevention. Future work will focus on refining accuracy and expanding the dataset.

\ifreview

\else
    \section{Acknowledgement}
    Supported by JSTSPRING Grant Number JPMJSP2125.
\fi
{
    \small
    \bibliographystyle{ieeenat_fullname}
    \bibliography{main}

\begin{thebibliography}{87}
\providecommand{\natexlab}[1]{#1}
\providecommand{\url}[1]{\texttt{#1}}
\expandafter\ifx\csname urlstyle\endcsname\relax
  \providecommand{\doi}[1]{doi: #1}\else
  \providecommand{\doi}{doi: \begingroup \urlstyle{rm}\Url}\fi

\bibitem[{\'A}lvarez-Aparicio et~al.(2022){\'A}lvarez-Aparicio, Guerrero-Higueras, Gonz{\'a}lez-Santamarta, Campazas-Vega, Matell{\'a}n, and Fern{\'a}ndez-Llamas]{alvarez2022biometric}
Claudia {\'A}lvarez-Aparicio, {\'A}ngel~Manuel Guerrero-Higueras, Miguel~{\'A}ngel Gonz{\'a}lez-Santamarta, Adri{\'a}n Campazas-Vega, Vicente Matell{\'a}n, and Camino Fern{\'a}ndez-Llamas.
\newblock Biometric recognition through gait analysis.
\newblock \emph{Scientific Reports}, 12\penalty0 (1):\penalty0 14530, 2022.

\bibitem[Cai et~al.(2019)Cai, Ge, Liu, Cai, Cham, Yuan, and Thalmann]{cai2019exploiting}
Yujun Cai, Liuhao Ge, Jun Liu, Jianfei Cai, Tat-Jen Cham, Junsong Yuan, and Nadia~Magnenat Thalmann.
\newblock Exploiting spatial-temporal relationships for 3d pose estimation via graph convolutional networks.
\newblock In \emph{Proceedings of the IEEE/CVF international conference on computer vision}, pages 2272--2281, 2019.

\bibitem[Cao et~al.(2017)Cao, Simon, Wei, and Sheikh]{cao2017realtime}
Zhe Cao, Tomas Simon, Shih-En Wei, and Yaser Sheikh.
\newblock Realtime multi-person 2d pose estimation using part affinity fields.
\newblock In \emph{Proceedings of the IEEE conference on computer vision and pattern recognition}, pages 7291--7299, 2017.

\bibitem[{Cao} et~al.(2019){Cao}, {Hidalgo Martinez}, {Simon}, {Wei}, and {Sheikh}]{8765346openpose}
Z. {Cao}, G. {Hidalgo Martinez}, T. {Simon}, S. {Wei}, and Y.~A. {Sheikh}.
\newblock Openpose: Realtime multi-person 2d pose estimation using part affinity fields.
\newblock \emph{IEEE Transactions on Pattern Analysis and Machine Intelligence}, 2019.

\bibitem[Ceriola et~al.(2023)Ceriola, Mileti, Donati, and Patan{\`e}]{ceriola2023comparison}
Luca Ceriola, Ilaria Mileti, Marco Donati, and Fabrizio Patan{\`e}.
\newblock Comparison of video-based algorithms for 2d human kinematics estimation: A preliminary study.
\newblock In \emph{Journal of Physics: Conference Series}, page 012002. IOP Publishing, 2023.

\bibitem[Cheng et~al.(2020{\natexlab{a}})Cheng, Xiao, Wang, Shi, Huang, and Zhang]{cheng2020higherhrnet}
Bowen Cheng, Bin Xiao, Jingdong Wang, Honghui Shi, Thomas~S Huang, and Lei Zhang.
\newblock Higherhrnet: Scale-aware representation learning for bottom-up human pose estimation.
\newblock In \emph{Proceedings of the IEEE/CVF conference on computer vision and pattern recognition}, pages 5386--5395, 2020{\natexlab{a}}.

\bibitem[Cheng et~al.(2020{\natexlab{b}})Cheng, Yang, Wang, and Tan]{cheng20203d}
Yu Cheng, Bo Yang, Bo Wang, and Robby~T Tan.
\newblock 3d human pose estimation using spatio-temporal networks with explicit occlusion training.
\newblock In \emph{Proceedings of the AAAI Conference on Artificial Intelligence}, pages 10631--10638, 2020{\natexlab{b}}.

\bibitem[Chun et~al.(2023)Chun, Park, and Chang]{chun2023learnable}
Sungho Chun, Sungbum Park, and Ju~Yong Chang.
\newblock Learnable human mesh triangulation for 3d human pose and shape estimation.
\newblock In \emph{Proceedings of the IEEE/CVF Winter Conference on Applications of Computer Vision}, pages 2850--2859, 2023.

\bibitem[Ci et~al.(2019)Ci, Wang, Ma, and Wang]{ci2019optimizing}
Hai Ci, Chunyu Wang, Xiaoxuan Ma, and Yizhou Wang.
\newblock Optimizing network structure for 3d human pose estimation.
\newblock In \emph{Proceedings of the IEEE/CVF international conference on computer vision}, pages 2262--2271, 2019.

\bibitem[Ci et~al.(2020)Ci, Ma, Wang, and Wang]{ci2020locally}
Hai Ci, Xiaoxuan Ma, Chunyu Wang, and Yizhou Wang.
\newblock Locally connected network for monocular 3d human pose estimation.
\newblock \emph{IEEE Transactions on Pattern Analysis and Machine Intelligence}, 44\penalty0 (3):\penalty0 1429--1442, 2020.

\bibitem[Ci et~al.(2023)Ci, Wu, Zhu, Ma, Dong, Zhong, and Wang]{ci2023gfpose}
Hai Ci, Mingdong Wu, Wentao Zhu, Xiaoxuan Ma, Hao Dong, Fangwei Zhong, and Yizhou Wang.
\newblock Gfpose: Learning 3d human pose prior with gradient fields.
\newblock In \emph{Proceedings of the IEEE/CVF conference on computer vision and pattern recognition}, pages 4800--4810, 2023.

\bibitem[Dosovitskiy et~al.(2021)Dosovitskiy, Beyer, Kolesnikov, Weissenborn, Zhai, Unterthiner, Dehghani, Minderer, Heigold, Gelly, Uszkoreit, and Houlsby]{dosovitskiy2020image}
Alexey Dosovitskiy, Lucas Beyer, Alexander Kolesnikov, Dirk Weissenborn, Xiaohua Zhai, Thomas Unterthiner, Mostafa Dehghani, Matthias Minderer, Georg Heigold, Sylvain Gelly, Jakob Uszkoreit, and Neil Houlsby.
\newblock An image is worth 16x16 words: Transformers for image recognition at scale.
\newblock \emph{ICLR}, 2021.

\bibitem[Einfalt et~al.(2023)Einfalt, Ludwig, and Lienhart]{einfalt2023uplift}
Moritz Einfalt, Katja Ludwig, and Rainer Lienhart.
\newblock Uplift and upsample: Efficient 3d human pose estimation with uplifting transformers.
\newblock In \emph{Proceedings of the IEEE/CVF winter conference on applications of computer vision}, pages 2903--2913, 2023.

\bibitem[Endo et~al.(2022)Endo, Poston, Sullivan, Fei-Fei, Pohl, and Adeli]{endo2022gaitforemer}
Mark Endo, Kathleen~L Poston, Edith~V Sullivan, Li Fei-Fei, Kilian~M Pohl, and Ehsan Adeli.
\newblock Gaitforemer: Self-supervised pre-training of transformers via human motion forecasting for few-shot gait impairment severity estimation.
\newblock In \emph{International Conference on Medical Image Computing and Computer-Assisted Intervention}, pages 130--139. Springer, 2022.

\bibitem[Fang et~al.(2024)Fang, Yeung, and Fujii]{fang2024foul}
Jiale Fang, Calvin Yeung, and Keisuke Fujii.
\newblock Foul prediction with estimated poses from soccer broadcast video.
\newblock \emph{arXiv preprint arXiv:2402.09650}, 2024.

\bibitem[Fukushima et~al.(2024)Fukushima, Blauberger, Guedes~Russomanno, and Lames]{fukushima2024potential}
Takashi Fukushima, Patrick Blauberger, Tiago Guedes~Russomanno, and Martin Lames.
\newblock The potential of human pose estimation for motion capture in sports: a validation study.
\newblock \emph{Sports Engineering}, 27\penalty0 (1):\penalty0 19, 2024.

\bibitem[Geng et~al.(2021)Geng, Sun, Xiao, Zhang, and Wang]{geng2021bottom}
Zigang Geng, Ke Sun, Bin Xiao, Zhaoxiang Zhang, and Jingdong Wang.
\newblock Bottom-up human pose estimation via disentangled keypoint regression.
\newblock In \emph{Proceedings of the IEEE/CVF conference on computer vision and pattern recognition}, pages 14676--14686, 2021.

\bibitem[Haberkamp et~al.(2022)Haberkamp, Garcia, and Bazett-Jones]{haberkamp2022validity}
Lucas~D Haberkamp, Micah~C Garcia, and David~M Bazett-Jones.
\newblock Validity of an artificial intelligence, human pose estimation model for measuring single-leg squat kinematics.
\newblock \emph{Journal of Biomechanics}, 144:\penalty0 111333, 2022.

\bibitem[Hamilton et~al.(2024)Hamilton, Glavcheva-Laleva, Milon, Anil, Williams, Bishop, and Holt]{hamilton2024comparison}
Rebecca~I Hamilton, Zornitza Glavcheva-Laleva, Md~Imdadul~Haque Milon, Yeshwin Anil, Jenny Williams, Peter Bishop, and Catherine Holt.
\newblock Comparison of computational pose estimation models for joint angles with 3d motion capture.
\newblock \emph{Journal of Bodywork and Movement Therapies}, 40:\penalty0 315--319, 2024.

\bibitem[Ingwersen et~al.(2023)Ingwersen, Mikkelstrup, Jensen, Hannemose, and Dahl]{SportsPose}
Christian~Keilstrup Ingwersen, Christian Mikkelstrup, Janus~N{\o}rtoft Jensen, Morten~Rieger Hannemose, and Anders~Bjorholm Dahl.
\newblock Sportspose: A dynamic 3d sports pose dataset.
\newblock In \emph{Proceedings of the IEEE/CVF International Workshop on Computer Vision in Sports}, 2023.

\bibitem[Ionescu et~al.(2013)Ionescu, Papava, Olaru, and Sminchisescu]{h36m}
Catalin Ionescu, Dragos Papava, Vlad Olaru, and Cristian Sminchisescu.
\newblock Human3. 6m: Large scale datasets and predictive methods for 3d human sensing in natural environments.
\newblock \emph{IEEE transactions on pattern analysis and machine intelligence}, 36\penalty0 (7):\penalty0 1325--1339, 2013.

\bibitem[Iskakov et~al.(2019)Iskakov, Burkov, Lempitsky, and Malkov]{iskakov2019learnable}
Karim Iskakov, Egor Burkov, Victor Lempitsky, and Yury Malkov.
\newblock Learnable triangulation of human pose.
\newblock In \emph{Proceedings of the IEEE/CVF international conference on computer vision}, pages 7718--7727, 2019.

\bibitem[Jiang et~al.(2023)Jiang, Lu, Zhang, Ma, Han, Lyu, Li, and Chen]{jiang2023rtmpose}
Tao Jiang, Peng Lu, Li Zhang, Ningsheng Ma, Rui Han, Chengqi Lyu, Yining Li, and Kai Chen.
\newblock Rtmpose: Real-time multi-person pose estimation based on mmpose.
\newblock \emph{arXiv preprint arXiv:2303.07399}, 2023.

\bibitem[Jiang et~al.(2024)Jiang, Billingham, M{\"u}ksch, Zarate, Evans, Oswald, Polleyfeys, Hilliges, Kaufmann, and Song]{jiang2024worldpose}
Tianjian Jiang, Johsan Billingham, Sebastian M{\"u}ksch, Juan Zarate, Nicolas Evans, Martin~R Oswald, Marc Polleyfeys, Otmar Hilliges, Manuel Kaufmann, and Jie Song.
\newblock Worldpose: A world cup dataset for global 3d human pose estimation.
\newblock In \emph{European Conference on Computer Vision}, pages 343--362. Springer, 2024.

\bibitem[Joo et~al.(2015)Joo, Liu, Tan, Gui, Nabbe, Matthews, Kanade, Nobuhara, and Sheikh]{panoptic}
Hanbyul Joo, Hao Liu, Lei Tan, Lin Gui, Bart Nabbe, Iain Matthews, Takeo Kanade, Shohei Nobuhara, and Yaser Sheikh.
\newblock Panoptic studio: A massively multiview system for social motion capture.
\newblock In \emph{Proceedings of the IEEE International Conference on Computer Vision}, pages 3334--3342, 2015.

\bibitem[Kr{\"u}ger et~al.(2016)Kr{\"u}ger, V{\"o}gele, Willig, Yao, Klein, and Weber]{kruger2016efficient}
Bj{\"o}rn Kr{\"u}ger, Anna V{\"o}gele, Tobias Willig, Angela Yao, Reinhard Klein, and Andreas Weber.
\newblock Efficient unsupervised temporal segmentation of motion data.
\newblock \emph{IEEE Transactions on Multimedia}, 19\penalty0 (4):\penalty0 797--812, 2016.

\bibitem[Lehrmann et~al.(2014)Lehrmann, Gehler, and Nowozin]{lehrmann2014efficient}
Andreas~M Lehrmann, Peter~V Gehler, and Sebastian Nowozin.
\newblock Efficient nonlinear markov models for human motion.
\newblock In \emph{Proceedings of the IEEE conference on computer vision and pattern recognition}, pages 1314--1321, 2014.

\bibitem[Li et~al.(2021{\natexlab{a}})Li, Wang, Zhang, Xu, Xu, and Tu]{li2021pose}
Ke Li, Shijie Wang, Xiang Zhang, Yifan Xu, Weijian Xu, and Zhuowen Tu.
\newblock Pose recognition with cascade transformers.
\newblock In \emph{Proceedings of the IEEE/CVF conference on computer vision and pattern recognition}, pages 1944--1953, 2021{\natexlab{a}}.

\bibitem[Li et~al.(2022{\natexlab{a}})Li, Wang, Gao, Song, Liu, Li, and Qiao]{iclr2022UniFormer}
Kunchang Li, Yali Wang, Peng Gao, Guanglu Song, Yu Liu, Hongsheng Li, and Yu Qiao.
\newblock Uniformer: Unified transformer for efficient spatiotemporal representation learning.
\newblock In \emph{International Conference on Learning Representations}, 2022{\natexlab{a}}.

\bibitem[Li et~al.(2021{\natexlab{b}})Li, Wang, Ni, Wang, Yang, and Zhang]{li20213d}
Linguo Li, Minsi Wang, Bingbing Ni, Hang Wang, Jiancheng Yang, and Wenjun Zhang.
\newblock 3d human action representation learning via cross-view consistency pursuit.
\newblock In \emph{Proceedings of the IEEE/CVF conference on computer vision and pattern recognition}, pages 4741--4750, 2021{\natexlab{b}}.

\bibitem[Li et~al.(2021{\natexlab{c}})Li, Yang, Ross, and Kanazawa]{AIST++}
Ruilong Li, Sha Yang, David~A. Ross, and Angjoo Kanazawa.
\newblock Ai choreographer: Music conditioned 3d dance generation with aist++.
\newblock \emph{2021 IEEE/CVF International Conference on Computer Vision (ICCV)}, pages 13381--13392, 2021{\natexlab{c}}.

\bibitem[Li et~al.(2024)Li, Wang, Liu, Tan, Lin, Wu, Chen, Zheng, and Li]{iclr2024MogaNet}
Siyuan Li, Zedong Wang, Zicheng Liu, Cheng Tan, Haitao Lin, Di Wu, Zhiyuan Chen, Jiangbin Zheng, and Stan~Z. Li.
\newblock Moganet: Multi-order gated aggregation network.
\newblock In \emph{International Conference on Learning Representations}, 2024.

\bibitem[Li et~al.(2022{\natexlab{b}})Li, Liu, Ding, Liu, Wang, and Yang]{li2022exploiting}
Wenhao Li, Hong Liu, Runwei Ding, Mengyuan Liu, Pichao Wang, and Wenming Yang.
\newblock Exploiting temporal contexts with strided transformer for 3d human pose estimation.
\newblock \emph{IEEE Transactions on Multimedia}, 25:\penalty0 1282--1293, 2022{\natexlab{b}}.

\bibitem[Li et~al.(2022{\natexlab{c}})Li, Liu, Tang, Wang, and Van~Gool]{li2022mhformer}
Wenhao Li, Hong Liu, Hao Tang, Pichao Wang, and Luc Van~Gool.
\newblock Mhformer: Multi-hypothesis transformer for 3d human pose estimation.
\newblock In \emph{Proceedings of the IEEE/CVF conference on computer vision and pattern recognition}, pages 13147--13156, 2022{\natexlab{c}}.

\bibitem[Li et~al.(2021{\natexlab{d}})Li, Zhang, Wang, Yang, Yang, Xia, and Zhou]{li2021tokenpose}
Yanjie Li, Shoukui Zhang, Zhicheng Wang, Sen Yang, Wankou Yang, Shu-Tao Xia, and Erjin Zhou.
\newblock Tokenpose: Learning keypoint tokens for human pose estimation.
\newblock In \emph{Proceedings of the IEEE/CVF International conference on computer vision}, pages 11313--11322, 2021{\natexlab{d}}.

\bibitem[Lin et~al.(2014)Lin, Maire, Belongie, Hays, Perona, Ramanan, Doll{\'a}r, and Zitnick]{cocodataset}
Tsung-Yi Lin, Michael Maire, Serge Belongie, James Hays, Pietro Perona, Deva Ramanan, Piotr Doll{\'a}r, and C~Lawrence Zitnick.
\newblock Microsoft coco: Common objects in context.
\newblock In \emph{Computer Vision--ECCV 2014: 13th European Conference, Zurich, Switzerland, September 6-12, 2014, Proceedings, Part V 13}, pages 740--755. Springer, 2014.

\bibitem[Liu et~al.(2025)Liu, Liu, Liu, and Li]{liu2025tcpformer}
Jiajie Liu, Mengyuan Liu, Hong Liu, and Wenhao Li.
\newblock Tcpformer: Learning temporal correlation with implicit pose proxy for 3d human pose estimation.
\newblock \emph{arXiv preprint arXiv:2501.01770}, 2025.

\bibitem[Liu et~al.(2022{\natexlab{a}})Liu, Chen, Chen, Chen, Xiao, Wu, K{\"a}rkk{\"a}inen, Pechenizkiy, Mocanu, and Wang]{liu2022more}
Shiwei Liu, Tianlong Chen, Xiaohan Chen, Xuxi Chen, Qiao Xiao, Boqian Wu, Tommi K{\"a}rkk{\"a}inen, Mykola Pechenizkiy, Decebal Mocanu, and Zhangyang Wang.
\newblock More convnets in the 2020s: Scaling up kernels beyond 51x51 using sparsity.
\newblock \emph{arXiv preprint arXiv:2207.03620}, 2022{\natexlab{a}}.

\bibitem[Liu et~al.(2021)Liu, Lin, Cao, Hu, Wei, Zhang, Lin, and Guo]{liu2021swin}
Ze Liu, Yutong Lin, Yue Cao, Han Hu, Yixuan Wei, Zheng Zhang, Stephen Lin, and Baining Guo.
\newblock Swin transformer: Hierarchical vision transformer using shifted windows.
\newblock In \emph{Proceedings of the IEEE/CVF international conference on computer vision}, pages 10012--10022, 2021.

\bibitem[Liu et~al.(2022{\natexlab{b}})Liu, Mao, Wu, Feichtenhofer, Darrell, and Xie]{liu2022convnet}
Zhuang Liu, Hanzi Mao, Chao-Yuan Wu, Christoph Feichtenhofer, Trevor Darrell, and Saining Xie.
\newblock A convnet for the 2020s.
\newblock In \emph{Proceedings of the IEEE/CVF conference on computer vision and pattern recognition}, pages 11976--11986, 2022{\natexlab{b}}.

\bibitem[Lugaresi et~al.(2019)Lugaresi, Tang, Nash, McClanahan, Uboweja, Hays, Zhang, Chang, Yong, Lee, et~al.]{lugaresi2019mediapipe}
Camillo Lugaresi, Jiuqiang Tang, Hadon Nash, Chris McClanahan, Esha Uboweja, Michael Hays, Fan Zhang, Chuo-Ling Chang, Ming~Guang Yong, Juhyun Lee, et~al.
\newblock Mediapipe: A framework for building perception pipelines.
\newblock \emph{arXiv preprint arXiv:1906.08172}, 2019.

\bibitem[Makino and Tauchi(2022)]{makino2022kinematic}
Mizuki Makino and Kenji Tauchi.
\newblock Kinematic factors related to forward and vertical release velocity in male javelin throwers.
\newblock \emph{International Journal of Sport and Health Science}, 20:\penalty0 249--259, 2022.

\bibitem[Martinez et~al.(2017)Martinez, Hossain, Romero, and Little]{martinez2017simple}
Julieta Martinez, Rayat Hossain, Javier Romero, and James~J Little.
\newblock A simple yet effective baseline for 3d human pose estimation.
\newblock In \emph{Proceedings of the IEEE international conference on computer vision}, pages 2640--2649, 2017.

\bibitem[Mehraban et~al.(2024)Mehraban, Adeli, and Taati]{mehraban2024motionagformer}
Soroush Mehraban, Vida Adeli, and Babak Taati.
\newblock Motionagformer: Enhancing 3d human pose estimation with a transformer-gcnformer network.
\newblock In \emph{Proceedings of the IEEE/CVF winter conference on applications of computer vision}, pages 6920--6930, 2024.

\bibitem[Mehta et~al.(2017)Mehta, Rhodin, Casas, Fua, Sotnychenko, Xu, and Theobalt]{3dhp}
Dushyant Mehta, Helge Rhodin, Dan Casas, Pascal Fua, Oleksandr Sotnychenko, Weipeng Xu, and Christian Theobalt.
\newblock Monocular 3d human pose estimation in the wild using improved cnn supervision.
\newblock In \emph{3D Vision (3DV), 2017 Fifth International Conference on}. IEEE, 2017.

\bibitem[Menychtas et~al.(2023)Menychtas, Petrou, Kansizoglou, Giannakou, Grekidis, Gasteratos, Gourgoulis, Douda, Smilios, Michalopoulou, et~al.]{menychtas2023gait}
Dimitrios Menychtas, Nikolaos Petrou, Ioannis Kansizoglou, Erasmia Giannakou, Athanasios Grekidis, Antonios Gasteratos, Vassilios Gourgoulis, Eleni Douda, Ilias Smilios, Maria Michalopoulou, et~al.
\newblock Gait analysis comparison between manual marking, 2d pose estimation algorithms, and 3d marker-based system.
\newblock \emph{Frontiers in Rehabilitation Sciences}, 4:\penalty0 1238134, 2023.

\bibitem[Mirek et~al.(2007)Mirek, Rudzi{\'n}ska, and Szczudlik]{mirek2007assessment}
Elzbieta Mirek, Monika Rudzi{\'n}ska, and Andrzej Szczudlik.
\newblock The assessment of gait disorders in patients with parkinson's disease using the three-dimensional motion analysis system vicon.
\newblock \emph{Neurologia i neurochirurgia polska}, 41\penalty0 (2):\penalty0 128--133, 2007.

\bibitem[Nekoui and Cheng(2021)]{nekoui2021enhancing}
Mahdiar Nekoui and Li Cheng.
\newblock Enhancing human motion assessment by self-supervised representation learning.
\newblock In \emph{BMVC}, page 322, 2021.

\bibitem[Nibali et~al.(2021)Nibali, Millward, He, and Morgan]{ASPset}
Aiden Nibali, Joshua Millward, Zhen He, and Stuart Morgan.
\newblock {ASPset}: An outdoor sports pose video dataset with {3D} keypoint annotations.
\newblock \emph{Image and Vision Computing}, page 104196, 2021.

\bibitem[Pavllo et~al.(2019)Pavllo, Feichtenhofer, Grangier, and Auli]{pavllo20193d}
Dario Pavllo, Christoph Feichtenhofer, David Grangier, and Michael Auli.
\newblock 3d human pose estimation in video with temporal convolutions and semi-supervised training.
\newblock In \emph{Proceedings of the IEEE/CVF conference on computer vision and pattern recognition}, pages 7753--7762, 2019.

\bibitem[Peng et~al.(2024)Peng, Zhou, and Mok]{peng2024ktpformer}
Jihua Peng, Yanghong Zhou, and PY Mok.
\newblock Ktpformer: Kinematics and trajectory prior knowledge-enhanced transformer for 3d human pose estimation.
\newblock In \emph{Proceedings of the IEEE/CVF Conference on Computer Vision and Pattern Recognition}, pages 1123--1132, 2024.

\bibitem[Qian et~al.(2023)Qian, Tang, Zhang, Han, Xiao, Huang, and Lin]{qian2023hstformer}
Xiaoye Qian, Youbao Tang, Ning Zhang, Mei Han, Jing Xiao, Ming-Chun Huang, and Ruei-Sung Lin.
\newblock Hstformer: Hierarchical spatial-temporal transformers for 3d human pose estimation.
\newblock \emph{arXiv preprint arXiv:2301.07322}, 2023.

\bibitem[Rao et~al.(2022)Rao, Zhao, Tang, Zhou, Lim, and Lu]{rao2022hornet}
Yongming Rao, Wenliang Zhao, Yansong Tang, Jie Zhou, Ser~Nam Lim, and Jiwen Lu.
\newblock Hornet: Efficient high-order spatial interactions with recursive gated convolutions.
\newblock \emph{Advances in Neural Information Processing Systems}, 35:\penalty0 10353--10366, 2022.

\bibitem[Reddy et~al.(2021)Reddy, Guigues, Pishchulin, Eledath, and Narasimhan]{reddy2021tessetrack}
N~Dinesh Reddy, Laurent Guigues, Leonid Pishchulin, Jayan Eledath, and Srinivasa~G Narasimhan.
\newblock Tessetrack: End-to-end learnable multi-person articulated 3d pose tracking.
\newblock In \emph{Proceedings of the IEEE/CVF Conference on Computer Vision and Pattern Recognition}, pages 15190--15200, 2021.

\bibitem[Sandbakk et~al.(2012)Sandbakk, Ettema, and Holmberg]{sandbakk2012influence}
{\O}yvind Sandbakk, Gertjan Ettema, and Hans-Christer Holmberg.
\newblock The influence of incline and speed on work rate, gross efficiency and kinematics of roller ski skating.
\newblock \emph{European journal of applied physiology}, 112:\penalty0 2829--2838, 2012.

\bibitem[Shan et~al.(2022)Shan, Liu, Zhang, Wang, Ma, and Gao]{shan2022p}
Wenkang Shan, Zhenhua Liu, Xinfeng Zhang, Shanshe Wang, Siwei Ma, and Wen Gao.
\newblock P-stmo: Pre-trained spatial temporal many-to-one model for 3d human pose estimation.
\newblock In \emph{European Conference on Computer Vision}, pages 461--478. Springer, 2022.

\bibitem[Shan et~al.(2023)Shan, Liu, Zhang, Wang, Han, Wang, Ma, and Gao]{shan2023diffusion}
Wenkang Shan, Zhenhua Liu, Xinfeng Zhang, Zhao Wang, Kai Han, Shanshe Wang, Siwei Ma, and Wen Gao.
\newblock Diffusion-based 3d human pose estimation with multi-hypothesis aggregation.
\newblock In \emph{Proceedings of the IEEE/CVF International Conference on Computer Vision}, pages 14761--14771, 2023.

\bibitem[Sigal et~al.(2010)Sigal, Balan, and Black]{Humaneva}
Leonid Sigal, Alexandru~O Balan, and Michael~J Black.
\newblock Humaneva: Synchronized video and motion capture dataset and baseline algorithm for evaluation of articulated human motion.
\newblock \emph{International journal of computer vision}, 87\penalty0 (1):\penalty0 4--27, 2010.

\bibitem[Song et~al.(2001)Song, Goncalves, and Perona]{song2001unsupervised}
Yang Song, Luis Goncalves, and Pietro Perona.
\newblock Unsupervised learning of human motion models.
\newblock \emph{Advances in Neural Information Processing Systems}, 14, 2001.

\bibitem[Sun et~al.(2019)Sun, Xiao, Liu, and Wang]{sun2019deep}
Ke Sun, Bin Xiao, Dong Liu, and Jingdong Wang.
\newblock Deep high-resolution representation learning for human pose estimation.
\newblock In \emph{Proceedings of the IEEE/CVF conference on computer vision and pattern recognition}, pages 5693--5703, 2019.

\bibitem[Suzuki et~al.(2024)Suzuki, Tanaka, Takeda, and Fujii]{suzuki2024pseudo}
Tomohiro Suzuki, Ryota Tanaka, Kazuya Takeda, and Keisuke Fujii.
\newblock Pseudo-label based unsupervised fine-tuning of a monocular 3d pose estimation model for sports motions.
\newblock In \emph{Proceedings of the IEEE/CVF Conference on Computer Vision and Pattern Recognition}, pages 3315--3324, 2024.

\bibitem[Takigawa and Tauchi(2023)]{takigawa2023factors}
Hiroko Takigawa and Kenji Tauchi.
\newblock Factors in javelin throw that result in differences in throwing records between throwers with similar approach velocities.
\newblock \emph{International Journal of Sport and Health Science}, 21:\penalty0 153--159, 2023.

\bibitem[Tanaka et~al.(2024)Tanaka, Suzuki, and Fujii]{tanaka20243d}
Ryota Tanaka, Tomohiro Suzuki, and Keisuke Fujii.
\newblock 3d pose-based temporal action segmentation for figure skating: A fine-grained and jump procedure-aware annotation approach.
\newblock In \emph{Proceedings of the 7th ACM International Workshop on Multimedia Content Analysis in Sports}, pages 17--26, 2024.

\bibitem[Tang et~al.(2023)Tang, Qiu, Hao, Hong, and Yao]{tang20233d}
Zhenhua Tang, Zhaofan Qiu, Yanbin Hao, Richang Hong, and Ting Yao.
\newblock 3d human pose estimation with spatio-temporal criss-cross attention.
\newblock In \emph{Proceedings of the IEEE/CVF Conference on Computer Vision and Pattern Recognition}, pages 4790--4799, 2023.

\bibitem[Tompson et~al.(2014)Tompson, Jain, LeCun, and Bregler]{tompson2014joint}
Jonathan~J Tompson, Arjun Jain, Yann LeCun, and Christoph Bregler.
\newblock Joint training of a convolutional network and a graphical model for human pose estimation.
\newblock \emph{Advances in neural information processing systems}, 27, 2014.

\bibitem[Toshev and Szegedy(2014)]{toshev2014deeppose}
Alexander Toshev and Christian Szegedy.
\newblock Deeppose: Human pose estimation via deep neural networks.
\newblock In \emph{Proceedings of the IEEE conference on computer vision and pattern recognition}, pages 1653--1660, 2014.

\bibitem[Trabelsi et~al.(2013)Trabelsi, Mohammed, Chamroukhi, Oukhellou, and Amirat]{trabelsi2013unsupervised}
Dorra Trabelsi, Samer Mohammed, Faicel Chamroukhi, Latifa Oukhellou, and Yacine Amirat.
\newblock An unsupervised approach for automatic activity recognition based on hidden markov model regression.
\newblock \emph{IEEE Transactions on automation science and engineering}, 10\penalty0 (3):\penalty0 829--835, 2013.

\bibitem[Trumble et~al.(2017)Trumble, Gilbert, Malleson, Hilton, and Collomosse]{TotalCapture}
Matt Trumble, Andrew Gilbert, Charles Malleson, Adrian Hilton, and John Collomosse.
\newblock Total capture: 3d human pose estimation fusing video and inertial sensors.
\newblock In \emph{2017 British Machine Vision Conference (BMVC)}, 2017.

\bibitem[Vaswani et~al.(2017)Vaswani, Shazeer, Parmar, Uszkoreit, Jones, Gomez, Kaiser, and Polosukhin]{vaswani2017attention}
Ashish Vaswani, Noam Shazeer, Niki Parmar, Jakob Uszkoreit, Llion Jones, Aidan~N Gomez, {\L}ukasz Kaiser, and Illia Polosukhin.
\newblock Attention is all you need.
\newblock \emph{Advances in neural information processing systems}, 30, 2017.

\bibitem[Von~Marcard et~al.(2018)Von~Marcard, Henschel, Black, Rosenhahn, and Pons-Moll]{3dpw}
Timo Von~Marcard, Roberto Henschel, Michael~J Black, Bodo Rosenhahn, and Gerard Pons-Moll.
\newblock Recovering accurate 3d human pose in the wild using imus and a moving camera.
\newblock In \emph{Proceedings of the European conference on computer vision (ECCV)}, pages 601--617, 2018.

\bibitem[Wang et~al.(2020)Wang, Yan, Xiong, and Lin]{wang2020motion}
Jingbo Wang, Sijie Yan, Yuanjun Xiong, and Dahua Lin.
\newblock Motion guided 3d pose estimation from videos.
\newblock In \emph{European conference on computer vision}, pages 764--780. Springer, 2020.

\bibitem[Wang et~al.(2022)Wang, Wen, Si, Qian, and Wang]{wang2022contrast}
Peng Wang, Jun Wen, Chenyang Si, Yuntao Qian, and Liang Wang.
\newblock Contrast-reconstruction representation learning for self-supervised skeleton-based action recognition.
\newblock \emph{IEEE Transactions on Image Processing}, 31:\penalty0 6224--6238, 2022.

\bibitem[Washabaugh et~al.(2022)Washabaugh, Shanmugam, Ranganathan, and Krishnan]{washabaugh2022comparing}
Edward~P Washabaugh, Thanikai~Adhithiyan Shanmugam, Rajiv Ranganathan, and Chandramouli Krishnan.
\newblock Comparing the accuracy of open-source pose estimation methods for measuring gait kinematics.
\newblock \emph{Gait \& posture}, 97:\penalty0 188--195, 2022.

\bibitem[Xiao et~al.(2018)Xiao, Wu, and Wei]{xiao2018simple}
Bin Xiao, Haiping Wu, and Yichen Wei.
\newblock Simple baselines for human pose estimation and tracking.
\newblock In \emph{Proceedings of the European conference on computer vision (ECCV)}, pages 466--481, 2018.

\bibitem[Xu et~al.(2022)Xu, Zhang, Zhang, and Tao]{xu2022vitpose}
Yufei Xu, Jing Zhang, Qiming Zhang, and Dacheng Tao.
\newblock Vitpose: Simple vision transformer baselines for human pose estimation.
\newblock \emph{Advances in Neural Information Processing Systems}, 35:\penalty0 38571--38584, 2022.

\bibitem[Yang et~al.(2022)Yang, Li, Dai, and Gao]{yang2022focal}
Jianwei Yang, Chunyuan Li, Xiyang Dai, and Jianfeng Gao.
\newblock Focal modulation networks.
\newblock \emph{Advances in Neural Information Processing Systems}, 35:\penalty0 4203--4217, 2022.

\bibitem[Yang et~al.(2021)Yang, Quan, Nie, and Yang]{yang2021transpose}
Sen Yang, Zhibin Quan, Mu Nie, and Wankou Yang.
\newblock Transpose: Keypoint localization via transformer.
\newblock In \emph{Proceedings of the IEEE/CVF international conference on computer vision}, pages 11802--11812, 2021.

\bibitem[Yang et~al.(2020)Yang, Zhu, Wu, Qian, Zhou, Zhou, and Loy]{yang2020transmomo}
Zhuoqian Yang, Wentao Zhu, Wayne Wu, Chen Qian, Qiang Zhou, Bolei Zhou, and Chen~Change Loy.
\newblock Transmomo: Invariance-driven unsupervised video motion retargeting.
\newblock In \emph{Proceedings of the IEEE/CVF Conference on Computer Vision and Pattern Recognition}, pages 5306--5315, 2020.

\bibitem[Yang et~al.(2023)Yang, Zeng, Yuan, and Li]{yang2023effective}
Zhendong Yang, Ailing Zeng, Chun Yuan, and Yu Li.
\newblock Effective whole-body pose estimation with two-stages distillation.
\newblock In \emph{Proceedings of the IEEE/CVF International Conference on Computer Vision}, pages 4210--4220, 2023.

\bibitem[Yeung et~al.(2024)Yeung, Ide, and Fujii]{AutoSoccerPose}
Calvin Yeung, Kenjiro Ide, and Keisuke Fujii.
\newblock Autosoccerpose: Automated 3d posture analysis of soccer shot movements.
\newblock In \emph{Proceedings of the IEEE/CVF Conference on Computer Vision and Pattern Recognition (CVPR) Workshops}, pages 3214--3224, 2024.

\bibitem[Yuan et~al.(2021)Yuan, Fu, Huang, Lin, Zhang, Chen, and Wang]{yuan2021hrformer}
Yuhui Yuan, Rao Fu, Lang Huang, Weihong Lin, Chao Zhang, Xilin Chen, and Jingdong Wang.
\newblock Hrformer: High-resolution transformer for dense prediction.
\newblock \emph{arXiv preprint arXiv:2110.09408}, 2021.

\bibitem[Zhang et~al.(2022)Zhang, Tu, Yang, Chen, and Yuan]{zhang2022mixste}
Jinlu Zhang, Zhigang Tu, Jianyu Yang, Yujin Chen, and Junsong Yuan.
\newblock Mixste: Seq2seq mixed spatio-temporal encoder for 3d human pose estimation in video.
\newblock In \emph{Proceedings of the IEEE/CVF conference on computer vision and pattern recognition}, pages 13232--13242, 2022.

\bibitem[Zhang et~al.(2019)Zhang, Li, Dong, Rosin, Cai, Han, Yang, Huang, and Hu]{zhang2019pose2seg}
Song-Hai Zhang, Ruilong Li, Xin Dong, Paul Rosin, Zixi Cai, Xi Han, Dingcheng Yang, Haozhi Huang, and Shi-Min Hu.
\newblock Pose2seg: Detection free human instance segmentation.
\newblock In \emph{Proceedings of the IEEE/CVF conference on computer vision and pattern recognition}, pages 889--898, 2019.

\bibitem[Zhao et~al.(2023)Zhao, Zheng, Liu, Wang, and Chen]{zhao2023poseformerv2}
Qitao Zhao, Ce Zheng, Mengyuan Liu, Pichao Wang, and Chen Chen.
\newblock Poseformerv2: Exploring frequency domain for efficient and robust 3d human pose estimation.
\newblock In \emph{Proceedings of the IEEE/CVF conference on computer vision and pattern recognition}, pages 8877--8886, 2023.

\bibitem[Zheng et~al.(2021)Zheng, Zhu, Mendieta, Yang, Chen, and Ding]{zheng20213d}
Ce Zheng, Sijie Zhu, Matias Mendieta, Taojiannan Yang, Chen Chen, and Zhengming Ding.
\newblock 3d human pose estimation with spatial and temporal transformers.
\newblock In \emph{Proceedings of the IEEE/CVF international conference on computer vision}, pages 11656--11665, 2021.

\bibitem[Zhu et~al.(2022)Zhu, Yang, Di, Wu, Wang, and Loy]{zhu2022mocanet}
Wentao Zhu, Zhuoqian Yang, Ziang Di, Wayne Wu, Yizhou Wang, and Chen~Change Loy.
\newblock Mocanet: Motion retargeting in-the-wild via canonicalization networks.
\newblock In \emph{Proceedings of the AAAI Conference on Artificial Intelligence}, pages 3617--3625, 2022.

\bibitem[Zhu et~al.(2023)Zhu, Ma, Liu, Liu, Wu, and Wang]{motionbert2022}
Wentao Zhu, Xiaoxuan Ma, Zhaoyang Liu, Libin Liu, Wayne Wu, and Yizhou Wang.
\newblock Motionbert: A unified perspective on learning human motion representations.
\newblock In \emph{Proceedings of the IEEE/CVF International Conference on Computer Vision}, 2023.

\end{thebibliography}
}


\end{document}